\documentclass[10pt,twocolumn,letterpaper]{article}

\usepackage{wacv}              % To produce the CAMERA-READY version
\usepackage{graphicx}
\usepackage{amsmath}
\usepackage{amssymb}
\usepackage{booktabs}
\usepackage{pifont}
\usepackage{amsfonts}
\usepackage{multirow}
\usepackage{tabularx}
\usepackage[table]{xcolor}
\usepackage{hyperref}
\usepackage{algorithm,algorithmic}
\usepackage{xcolor}
\begin{document}

%%%%%%%%% TITLE - PLEASE UPDATE
\title{Wavelet-Driven Generalizable Framework for Deepfake Face Forgery Detection}

\author{Lalith Bharadwaj Baru\(^{1,*}\), Rohit Boddeda\(^{1,*}\), Shilhora Akshay Patel\(^{2,*}\), Sai Mohan Gajapaka\(^{3}\)\\
\(^{1}\)International Institute of Information Technology Hyderabad, TS, India.\\
\(^{2}\)Indian Institute of Technology Hyderabad, TS, India. \\
\(^{3}\)Michigan State University, MI, USA \\
Equal contribution as first authors.\\
\(\tt \small\) \{lalith.baru,rohit.b\}@research.iiit.ac.in\\}

\maketitle

%%%%%%%%% ABSTRACT
\begin{abstract}
   The evolution of digital image manipulation, particularly with the advancement of deep generative models, significantly challenges existing deepfake detection methods, especially when the origin of the deepfake is obscure. To tackle the increasing complexity of these forgeries, we propose \textbf{Wavelet-CLIP}, a deepfake detection framework that integrates wavelet transforms with features derived from the ViT-L/14 architecture, pre-trained in the CLIP fashion. Wavelet-CLIP utilizes Wavelet Transforms to deeply analyze both spatial and frequency features from images, thus enhancing the model's capability to detect sophisticated deepfakes. To verify the effectiveness of our approach, we conducted extensive evaluations against existing state-of-the-art methods for cross-dataset generalization and detection of unseen images generated by standard diffusion models. Our method showcases outstanding performance, achieving an average AUC of 0.749 for cross-data generalization and 0.893 for robustness against unseen deepfakes, outperforming all compared methods. The code can be reproduced from the repo: \url{https://github.com/lalithbharadwajbaru/wavelet-clip}
\end{abstract}

%%%%%%%%% BODY TEXT

\section{Introduction}
In today's digital landscape, we are witnessing an inundation of counterfeit images, arising from various sources. Some of these images are manipulated versions of authentic photos, altered using tools such as FaseShifter \cite{li2020fsh} and proprietor photoshop tools \cite{wang2019detecting}, while others are crafted through advanced machine learning algorithms. The advent and refinement of deep generative models \cite{ho2020denoising, song2020denoising, rombach2022high} have particularly highlighted the latter category, drawing both admiration for the photo-realistic images they can produce and concern over their potential misuse. The challenge is compounded by the diverse origins of these fake images; they may manifest as real human faces created by generative adversarial networks or as intricate scenes synthesized by diffusion models \cite{rombach2022high}. This growing variety underscores the inevitability of encountering new forms of image forgery. 

The proliferation of diffusion models \cite{ho2020denoising, song2020denoising, rombach2022high} has revolutionized the field of generative AI, enabling the creation of highly realistic synthetic images with exceptional quality and diversity. These models have demonstrated remarkable capabilities in producing photorealistic human faces, complex natural scenes, and seamlessly manipulated content. However, this rapid advancement has also raised significant concerns regarding their potential misuse for malicious purposes, such as generating deepfake media or spreading disinformation. As diffusion models continue to evolve, it becomes increasingly challenging to distinguish between real and fake images, amplifying the need for robust fake image detection systems that can generalize across diverse generative families. Against this challenges, our research aims to devise a diverse generalizable fake detection framework capable of identifying any falsified image, even when training is confined to a single type of generative model. 

Traditionally, fake image detection has been approached as a binary classification task, where deep neural networks are trained to distinguish real images from synthetic ones generated by a specific model, such as diffusion or GAN methods. While these approaches excel within the same generative family (e.g., detecting fake images produced by diffusion variants like LDM \cite{rombach2022high} or Guided Diffusion \cite{dhariwal2021diffusion}), they fail to generalize when exposed to unseen generative families. This limitation arises because these classifiers tend to rely on low-level artifacts unique to the training model, often referred to as "fingerprints." Consequently, fake images generated by alternative methods that lack these specific fingerprints are misclassified as real, leading to a skewed decision boundary and poor generalization to novel image generation techniques.

There are numerous methods developed for deepfake generalization both within and cross-domain evaluation \cite{yan2024deepfakebench, ni2022core, afchar2018mesonet, tan2019efficientnet, yan2023ucf}. Of which, some works rely of basic encoders such as EfficentNet \cite{tan2019efficientnet} and Xception \cite{chollet2017xception}. Some models leverage frequency-based statistics for identfying some details which spatial domain can't capture \cite{qian2020thinking,liu2021spatial,luo2021generalizing}. Most of the existing deepfake detection models demonstrate significant results in scenarios where the training and testing data come from the same dataset. However, these detectors frequently face challenges in cross-domain or cross-dataset scenarios, where there is a significant discrepancy between the distribution of the training data and that of the testing data.

Our method significantly advances the field of digital forensics by offering a robust model capable of countering the evolving threat of digital image forgery. To address these limitations,we avoid explicit learning for real-vs-fake image classification and utilize the feature space of a large pre-trained vision-language model, CLIP-ViT \cite{radford2021clip} which has been trained on internet-scale datasets for tasks unrelated to fake detection. It achieves this by effectively generalizing across different datasets and adeptly identifying deepfakes produced by powerful, previously unseen generators. The proposed approach offers distinct advantages over current methodologies in two key directions:

\begin{enumerate}
    \item We introduce an innovative Wavelet-based classifier designed specifically for deepfake detection, showcasing its applicability in identifying manipulated (deepfake) content.
    \item Next, we highlight the capability of representations derived from CLIP to not only perform effectively across different unseen datasets but also to accurately identify images generated by models trained on previously unseen datasets.
\end{enumerate}

\begin{figure*}[!t]
    \centering
    \includegraphics[width=1.02\textwidth]{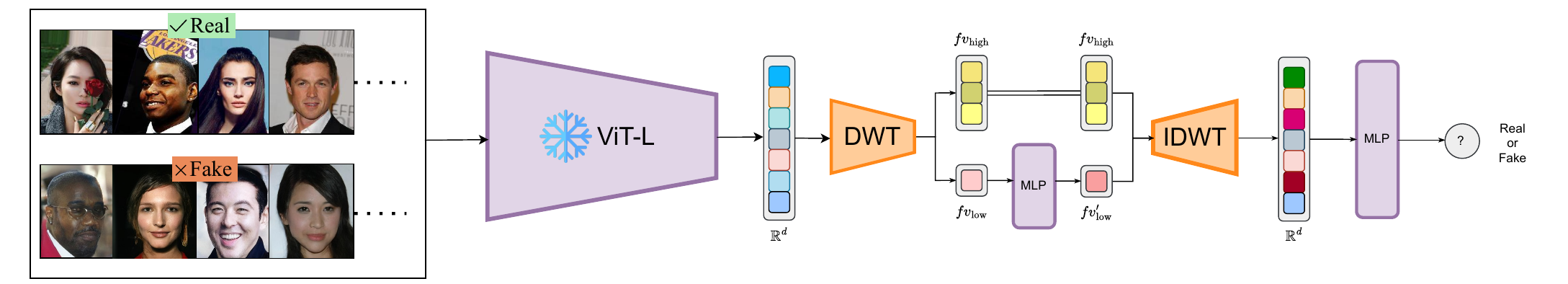}
    \caption{\textbf{Wavelet-CLIP:} The comprehensive workflow of the proposed Wavelet-CLIP. Initially, the model ingests real and counterfeit image samples, which are processed by a ViT-L/14 encoder, pretrained with CLIP weights \cite{radford2021clip}, to produce feature representations. These representations are then subjected to Discrete Wavelet Transform (DWT) to downsample into low-frequency and high-frequency components. The low-frequency component is further refined using a MLP keeping the high frequency features $fv_{\text{high}}$ constant (where, the "$=$" signifies an identity mapping). Subsequently, the transformed representations are processed by another MLP to classify the image is a deepfake or genuine.}
    \label{fig-main}
\end{figure*}

\section{Related Works}
\subsection{Naive Detectors}
Naive detectors utilize the existing state-of-the-art CNN architectures to directly classify images as real or fake. These models do not rely on any handcrafted layers or domain-specific knowledge but instead learn features directly from the data during training. MesoNet \cite{afchar2018mesonet} and MesoInception \cite{afchar2018mesonet}, which are lightweight networks optimized for efficiency and designed to capture mesoscopic features indicative of manipulations. More advanced naive detectors, such as Xception \cite{chollet2017xception} and EfficientNet-B4 \cite{tan2019efficientnet}, employ modern CNN architectures that are computationally efficient. 

Naive detectors are simple to implement and often achieve reasonable performance but may struggle with generalization across datasets and sophisticated forgeries due to their reliance on learned features without deeper semantic insights.

\subsection{Spatial Detectors}
Spatial detectors focus on analyzing the spatial domain of images, often employing advanced techniques to detect localized artifacts introduced during manipulations. Models like Capsule Networks \cite{nguyen2019capsule} leverage dynamic routing between capsules to model spatial hierarchies irrespective of rotations, while DSP-FWA \cite{li2018exposing} specializes in detecting warping artifacts that occur during face-swapping manipulations. Face X-ray \cite{li2020facexray} zeroes in on boundary artifacts between manipulated and non-manipulated regions, leveraging high-resolution features to isolate forgery artifacts. Models like CORE \cite{ni2022core} and UCF \cite{yan2023ucf} emphasize learning robust and consistent features, enhancing generalization to unseen forgeries. 

By focusing on spatial inconsistencies such as blending, texture mismatches, or altered facial regions, spatial detectors excel at identifying specific manipulations but may require more sophisticated pre-processing and training strategies. These methods might be robust to visually perceptually forgeries but, can't perceive hidden forgeries.  

\subsection{Frequency Detectors}
Frequency detectors analyze the frequency domain of images to detect subtle artifacts not visible in the spatial domain. These models address the limitations of spatial detectors by capturing inconsistencies in high-frequency components, noise patterns, and phase information. For instance, F3Net \cite{qian2020thinking} uses adaptive filters to mine forgery clues in the frequency domain, making it effective at identifying hidden noise introduced during manipulations. SRM \cite{luo2021generalizing} employs high-frequency filters to identify subtle pixel-level inconsistencies.

Frequency detectors are particularly robust and operate in the frequency domain to identify manipulations by detecting artifacts in noise patterns or phase spectra. This complements spatial detectors by providing insights into overlooked forgery artifacts.

\subsection{Generalizable Detectors}
In Yan \emph{et al.} \cite{yan2024deepfakebench} study, the methods focus on cross-domain generalization but, deepfakes can emerge from nowhere. Thus, a generalizable model should have the capability of identifying unseen fake or forgery images. To address this challenge Ojha \emph{et al.} \cite{ojha2023towards} provided a new direction of solving unseen deepfakes generated from diffusion and autoregressive methods. Unlike traditional classifiers trained explicitly for real-vs-fake classification, which fail to generalize to new generative model families, the proposed approach leverages feature spaces from large, pre-trained vision-language models such as CLIP \cite{radford2021clip}. Later, cozzolino \emph{et al. }\cite{cozzolino2024raising} have comprehensively analyzed the frozen CLIP features are performed exhaustive experiments on various unseen deepfakes and showcased the significance of frozen CLIP features for generalizable deepfake identification.

\section{Methodology}
The main objective of this work is to devise a generalizable deepfake identification model which has two significant properties. First, the model is required to capture low-frequency features with detailed granular representations. Second, these representations should be adept at discerning forgery-specific characteristics, determining their authenticity or counterfeit nature.

% Thus, we intend to develop a feature extractor (or encoder) which is capable of extracting granular features and design a classifier that could distinguish between a deepfake and a camera-captured image. Thus, we divide the model into two segments a)Encoder and b) Classifier Head.

Therefore, our objective is to engineer a feature extractor (or encoder) which is capable of extracting granular features, alongside crafting a classifier that can effectively differentiate between deepfake and authentic camera-captured images. Thus, we partition the entire model into two primary components: a) the Encoder and b) the Classification Head.

\subsection{Encoder}
A good encoder has to understand the crucial features from the image distribution and map them to the latent space. These latent features should carry the prominent features of the image. But, when it comes to generalization, the features have to be more relevant irrespective of trained or seen samples. In such scenarios, a model that is trained on internet-scale data in a self-supervised fashion should provide fine-grained features irrespective of the nature of the data. Hence, we adopt a pre-trained vision transformer \cite{dosovitskiy2020vit} model that is trained via. CLIP fashion\cite{radford2021clip} pertaining strong one-shot transferable features. This encoder maps an image into a representation space of feature dimension $d$ where $Enc_{\phi}: \mathbb{R}^{ 256 \times 256 \times 3} \to \mathbb{R}^d$ (we denote frozen encoder as $\phi$). The latent features $Z$ captured for our study are using ViT-L/14 \cite{radford2021clip} and is represented as,

\begin{equation}
    Z = Enc_{\phi}^{(\text{ViT})}(x), \;\;\;\;\;\; Z \in \mathbb{R}^{768}
\end{equation}

These acquired representations allows to have strong feature space as they have been learned in self-supervised contrastive fashion without task-oriented training. Our chosen ViT L/14 encoder is not trained nor fine-tuned for deepfake identifications. Thus our encoder stands out a step ahead from the models that were designed and trained in a supervised fashion for deepfake forgery identification \cite{afchar2018mesonet, nguyen2019capsule,dang2020detection, qian2020thinking, yan2023ucf}. Training in a supervised fashion (on FaceForensics++ dataset \cite{rossler2019faceforensics++} c23) may not help to generalize on samples that are photo-realistic deepfakes generated from state-of-the-art diffusion models \cite{ojha2023towards}. Thus, ViT L/14 encoder has the strong generalizable representations which maps a real or deepfake images into a latent space, the next crucial step is to classify them using a strong generalizable classifier.

% Ojha \emph{et al. }\cite{ojha2023towards} uses features from ViT L trained in CLIP fashion. Use a k-NN or train a Linear Classifier(LC)  to classify the generated images sampled from each of the generative model. Thus thier major limitation is that they allow training a LC for each generative  model and the classification head lacks the generalizability. 

\subsection{Classification Head}
The classification head is tasked with categorizing the features generated by our encoder. Drawing inspiration from frequency-based techniques like Fourier Transforms \cite{qian2020thinking}, we focus on extracting subtle forgery indicators from images. We have developed a frequency-based Wavelet Classification Head that processes the features $Z$ derived from CLIP to determine their authenticity. In the following sections, we will provide a primer for the Discrete Wavelet Transforms and their inversions, and explain how certain design decisions can enhance the effectiveness of the classifier to identify deepfakes.

\begin{algorithm}[t]
\caption{Wavelet-CLIP}\label{algo-main}
\begin{algorithmic}[1]
% \STATE \textbf{\underline{Training}}
\STATE \textbf{Input:} \textsc{Dataset} $\mathcal{D}$, \textsc{Encoder} $Enc_{\phi}^{(\text{ViT})}(.)$, $\epsilon$, $n$;
\FOR{$\textsc{iterations} = 1$ \TO $\epsilon$}
\FOR {$\textsc{batch} = n$}
\STATE  $Z^{(n)} = Enc_{\phi}^{(\text{ViT})}(x^{(n)})$
\STATE $fv_{\text{low}}^{(n)}, fv_{\text{high}}^{(n)} = \text{DWT}(Z^{(n)})$
\STATE  ${fv'}_{\text{low}}^{(n)} = \text{MLP}(fv_{\text{low}}^{(n)})$
\STATE $Z_{\text{new}}^{(n)} = \text{IDWT}([{fv'}_{\text{low}}^{(n)}, fv_{\text{high}}^{(n)}])$
\STATE  $\text{cls}_n = \text{MLP}(Z_{\text{new}}^{(n)})$
\ENDFOR
\ENDFOR
\STATE   \textbf{return} $\text{cls}_n$
\end{algorithmic}
\end{algorithm}

\paragraph*{\bf Wavelet Transforms}
Wavelet Transforms are used to analyze various frequency components of a signal and is particularly useful representations that have hierarchical or multi-scale structure \cite{mallat1989theory}. Applying a Discrete Wavelet Transform (DWT) the representation splits into low and high frequency components. Low-frequency components are responsible in capturing broad and nuanced features. Whereas high frequency components capture sharp features. 

% The DWT decomposes as
% Low-pass filter (\( l \)), that captures the low-frequency component. Similarly, captures the high-frequency component, resulting in a down-sampled signal \( d_1 \). These are denoted as, 
% \begin{align}
%    s_{1k} &= \sum_j l_{(j-2k)} s_j\\
%    d_{1k} &= \sum_j h_{(j-2k)} s_j
% \end{align}
% These operations can be expressed in matrix form, where \( L \) and \( H \) are matrices constructed from low-pass and high-pass filter coefficients, respectively, performing filtering and down-sampling in a single step through the operations \( s_1 = Ls \) and \( d_1 = Hs \). The Inverse Discrete Wavelet Transform  IDWT reconstructs the original signal \( s \) from its low-frequency (\( s_1 \)) and high-frequency (\( d_1 \)) components using:
% \begin{equation}
%     s_j = \sum_k (l_{j-2k} s_{1k} + h_{j-2k} d_{1k})
% \end{equation}

The Discrete Wavelet Transform (DWT) of a one-dimensional signal \( s = \{s_j\}_{j \in \mathbb{Z}} \) decomposes it into two components: a low-frequency approximation \( s_1 = \{s_{1k}\}_{k \in \mathbb{Z}} \) and a high-frequency detail \( d_1 = \{d_{1k}\}_{k \in \mathbb{Z}} \). These components are defined as,

\begin{equation}
    s_{1k} = \sum_{j \in \mathbb{Z}} l_{j-2k} s_j, \quad d_{1k} = \sum_{j \in \mathbb{Z}} h_{j-2k} s_j,
\end{equation}

where \( l = \{l_k\}_{k \in \mathbb{Z}} \) and \( h = \{h_k\}_{k \in \mathbb{Z}} \) represent the low-pass and high-pass filters, respectively, associated with an orthogonal wavelet. The Inverse Discrete Wavelet Transform (IDWT) allows the reconstruction of the original signal \( s = \{s_j\}_{j \in \mathbb{Z}} \) from its low-frequency approximation \( s_1 = \{s_{1k}\}_{k \in \mathbb{Z}} \) and high-frequency detail \( d_1 = \{d_{1k}\}_{k \in \mathbb{Z}} \). The reconstruction is performed as,

\begin{equation}
    s_j = \sum_{k \in \mathbb{Z}} \left( l_{j-2k} s_{1k} + h_{j-2k} d_{1k} \right),
\end{equation}

where \( l = \{l_k\}_{k \in \mathbb{Z}} \) and \( h = \{h_k\}_{k \in \mathbb{Z}} \) are the low-pass and high-pass filters associated with the orthogonal wavelet. In this equation, \( l_{j-2k} \) acts as an interpolation filter that reconstructs the low-frequency components,\( h_{j-2k} \) reconstructs the high-frequency details,
 the summation runs over all integers \( k \), ensuring that both low-frequency and high-frequency components contribute to the reconstructed signal \( s_j \).

These operations can be expressed in matrix form, where \( L \) and \( H \) are matrices constructed from low-pass and high-pass filter coefficients, respectively. For 2D signals like images, applying DWT along both dimensions results in four components--\( X_{ll},\) (low-frequency in both dimensions), \( X_{lh} \) (low-frequency row-wise, high-frequency column-wise), \( X_{hl} \) (high-frequency row-wise, low-frequency column-wise), and \( X_{hh}\) (high-frequency in both dimensions)—by applying matrices \( L \) and \( H \) across rows and columns, respectively. The DWT and IDWT can be expressed as,

\begin{align}
X_{ll} &= L X L^T,  \\
X_{lh} &= H X L^T,  \\
X_{hl} &= L X H^T,  \\
X_{hh} &= H X H^T, 
\end{align}
\begin{equation}
  X = L^T X_{ll} L + H^T X_{lh} L + L^T X_{hl} H + H^T X_{hh} H.   
\end{equation}

% As we have to find the counterfeit representations, spatial representations are not sufficient for classification. 
\begin{table*}[!t]
    \centering
    \label{tab:dataset_details}
    \begin{tabular}{l|c|c|l}
        \hline
        \textbf{Dataset Name}       & \textbf{Train/Test} & \textbf{No. of Samples} & \textbf{Generalization Evaluation}                        \\ 
        \hline
        \hline
        FaceForensics++ \cite{rossler2019faceforensics++} & Train          & 114884                   & -                \\ 
        \hline
        Celeb-DF v1 (CDFv1) \cite{li2020celeb}   & Test           & 3136                     & Cross-domain                 \\ 
        Celeb-DF v2 (CDFv2) \cite{li2020celeb}   & Test           & 16420                     & Cross-domain                 \\ 
        FaceShifter (Fsh) \cite{li2020fsh}       & Test           & 8958                    & Cross-domain                 \\ 
        \hline
        Diffusion Models  (DDPM, DDIM, LDM)          & Test           & 50,000                  & Novel Face Deepfake             \\

        \hline
    \end{tabular}
    \caption{\textbf{Dataset Information:} The table provides a description of the datasets employed in our study to evaluate Cross-Domain and Unseen Deepfake Generalization. }\label{data-tab}
\end{table*}

\paragraph*{\bf Wavelet Classifier}
Now, we apply these transformations to the features derived from our encoder for effective classification. It is well-established that low-frequency components contain valuable information within the acquired representations. Therefore, to capture the most significant representations, we opt to transform the low-frequency features obtained from the DWT using an MLP layer (ref eq (11)) i.e., $X_{ll}$. This method facilitates the learning of broad and granular invariances. Subsequently, IDWT is employed to reconstruct these features into the spatial domain (ref eq (12)).  The refined representations post-transformation are instrumental in discerning low-frequency components and spatial details, thereby strengthening our capability to differentiate between authentic and deepfake representations effectively (ref fig \ref{fig-main}). The mathematical formulation can be described as,

\begin{align}
fv_{\text{low}}^{(n)}, fv_{\text{high}}^{(n)} &= \text{DWT}(Z^{(n)}) \\
{fv'}_{\text{low}}^{(n)} &= \text{MLP}(fv_{\text{low}}^{(n)}) \\
Z_{\text{new}}^{(n)} &= \text{IDWT}([{fv'}_{\text{low}}^{(n)}, fv_{\text{high}}^{(n)}])
\end{align}

The algorithm for our proposed model is delineated in Algorithm \ref{algo-main}, positioning it as a versatile deepfake detection solution. Essentially, the model's efficacy lies in the robust learning capabilities of a strong encoder to classify representations accurately. We will next evaluate the effectiveness of this methodology by analyzing the performance of our proposed framework across diverse experimental conditions.

\begin{figure}[t!]
    \centering
    % First Subfigure
    \begin{subfigure}[b]{0.50\textwidth}
        \centering
        \includegraphics[width=\textwidth]{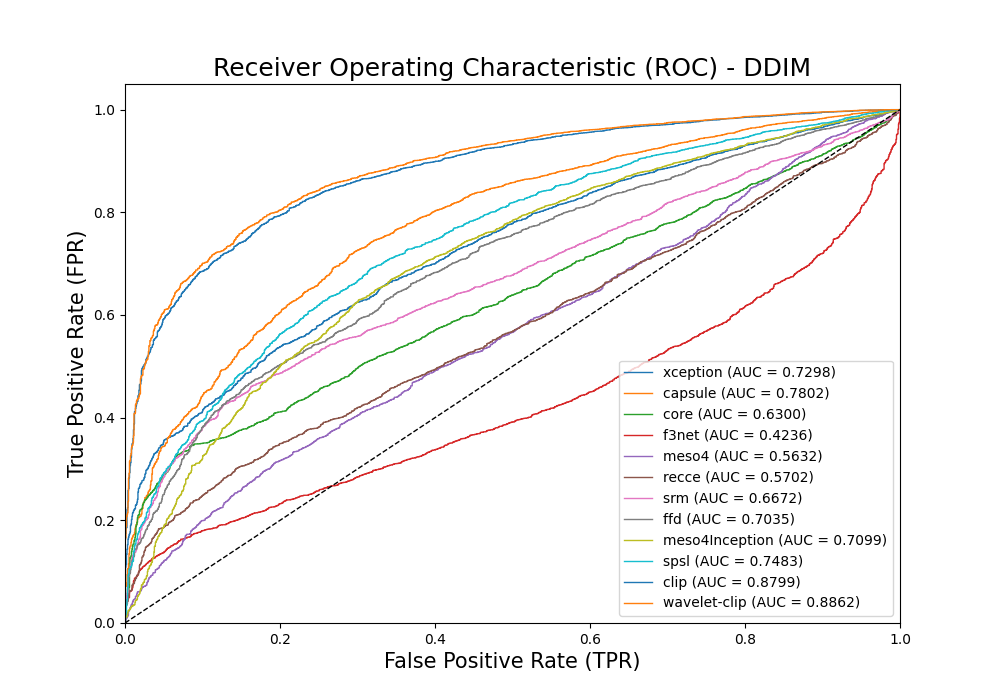}
        \caption{ROC - DDIM}
        \label{fig:roc_ddim}
    \end{subfigure}
    % Second Subfigure
    \begin{subfigure}[b]{0.50\textwidth}
        \centering
        \includegraphics[width=\textwidth]{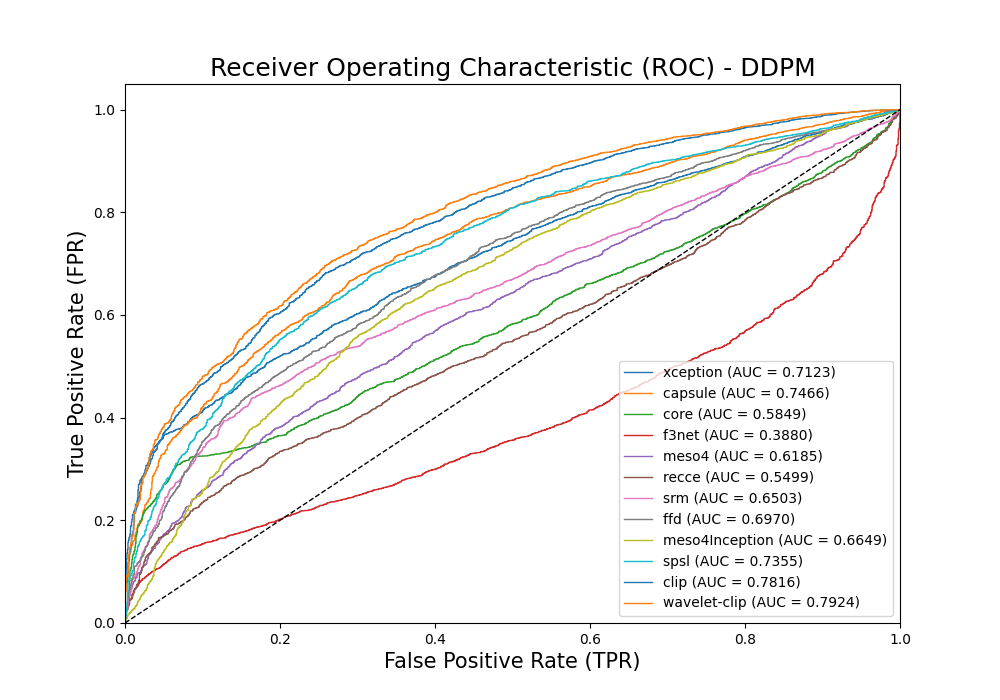}
        \caption{ROC - DDPM}
        \label{fig:roc_ddpm}
    \end{subfigure}
    % Third Subfigure
    \begin{subfigure}[b]{0.50\textwidth}
        \centering
        \includegraphics[width=\textwidth]{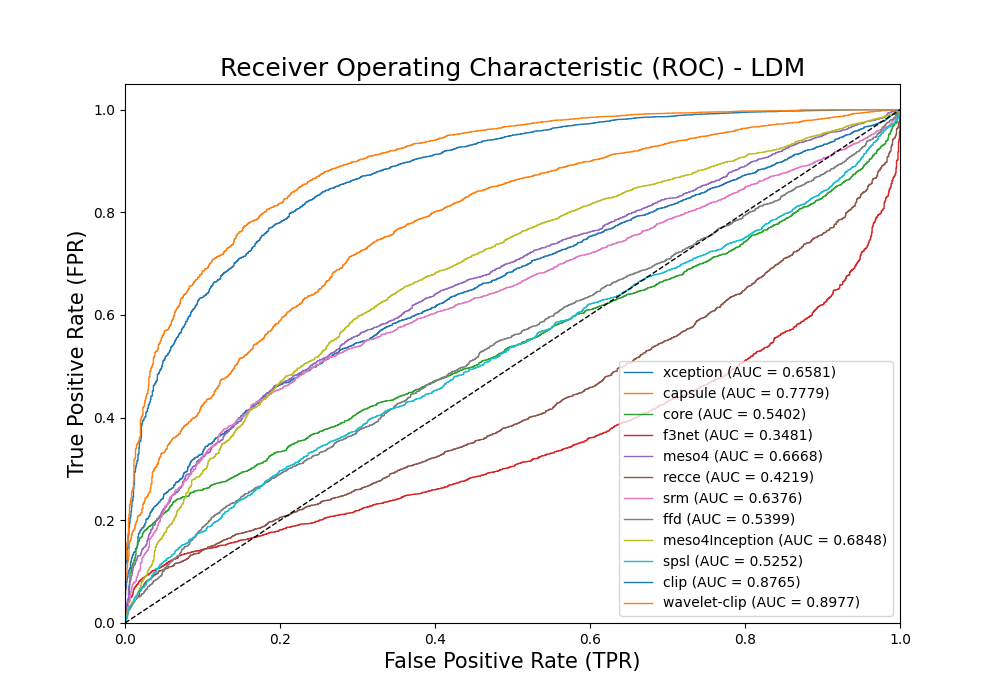}
        \caption{ROC - LDM}
        \label{fig:roc_ldm}
    \end{subfigure}
    \caption{\textbf{AUCROC Plots:} Receiver Operating Characteristic (ROC) curves for a) DDIM, b) DDPM, and c) LDM, depicting the models' performance in terms of the Area Under the Curve (AUC), along with their true positive and false positive rates.}
    \label{fig-roc-combined}
\end{figure}

\section{Set Up}
In this section, we will first detail the evaluation protocol, evaluation metrics, and datasets used.

\paragraph{Dataset and Evaluation} In alignment with the training and evaluation protocol established by Yan \emph{et al.} \cite{yan2024deepfakebench}, the models undergo initial training on the FaceForensics++ c23 dataset \cite{rossler2019faceforensics++}. Subsequent evaluations employ a cross-domain test using datasets such as Celeb-DF v1 (CDFv1) \cite{li2020celeb}, Celeb-DF v2 (CDFv2) \cite{li2020celeb}, and FaceShifter (Fsh) \cite{li2020fsh}, thereby providing a robust framework to test the generalization capabilities of all the models. While the existing benchmarks do not encompass tests on emerging diffusion models, our research extends to examining generalizability with novel synthetic samples. Utilizing state-of-the-art diffusion models like DDPM \cite{ho2020denoising}, DDIM \cite{song2020denoising}, and LDM \cite{rombach2022high} (wihtout text-guidance), we generate approximately 50,000 images from the CelebA dataset weights--none of which were included in  training phase. This approach enables a comprehensive evaluation of model’s adaptability to novel and unseen data, leveraging its potential for practical deployment in digital forensics (Refer to Table \ref{data-tab}). 

\paragraph{Metrics} To assess the effectiveness of the results we use AUC (Area Under the Curve) and EER (Equal Error Rate) as fundamental metrics \cite{yan2024deepfakebench}. The AUROC represents the degree of separability achieved by the model, indicating how well the model can distinguish between normal and anomalous images. The AUROC is calculated as the area under the ROC curve, which plots the true positive rate (TPR) against the false positive rate (FPR) at various threshold settings \cite{fawcett2006introduction}. The formula for AUROC can be expressed as,
\begin{equation}
    \text{AUROC} = \int_{0}^{1} TPR(FPR^{-1}(x)) dx
\end{equation}

The Equal Error Rate (EER) is a metric used to evaluate the performance of a binary classification system, especially in biometric verification or detection tasks. It is the point at which the False Acceptance Rate (FAR) and the False Rejection Rate (FRR) are equal. EER is defined as,
\begin{equation}
    \text{EER} = \text{FAR}(\tau^*) = \text{FRR}(\tau^*),
\end{equation}

where \( \tau^* \) is the decision threshold that minimizes the difference between FAR and FRR. The EER provides a single scalar value to compare systems: lower EER indicates better system performance. It is often visualized on a Receiver Operating Characteristic (ROC) curve as the point where the curve intersects the line \( \text{FPR} = 1 - \text{TPR} \).

\begin{table*}[h!]
\centering
\scalebox{0.99}{
\begin{tabular}{l|c|c|c|c|c|c|c}
\hline
\rowcolor{lightgray}\textbf{Models} & \textbf{Venue} & \textbf{Backbone} & \textbf{Protocol} & \textbf{CDFv1} & \textbf{CDFv2} & $\;\;$\textbf{Fsh} $\;\;$& \textbf{Avg.} \\
\hline

 MesoNet \cite{afchar2018mesonet}  & WIFS-18& Custom CNN & Supervised & 0.735 & 0.609 & 0.566 & 0.636 \\
 MesoInception \cite{afchar2018mesonet} & WIFS-18 & Inception &  Supervised & 0.736 & 0.696 & 0.643 & 0.692 \\
EfficentNet\cite{tan2019efficientnet}& ICML-19 & EfficentNet B4 \cite{tan2019efficientnet} & Supervised & 0.790 & 0.748 & 0.616  & 0.718 \\ \hline
Xception \cite{chollet2017xception} & ICCV-19   &  Xception &  Supervised&  0.779 & 0.736 & 0.624 & 0.713 \\
Capusle \cite{nguyen2019capsule} & ICASSP-19  & CapsuleNet \cite{sabour2017dynamic} & Supervised & 0.790 & 0.747 & 0.646 & 0.728 \\
DSP-FWA \cite{li2018exposing} & CVPR-19 & Xception \cite{chollet2017xception}  & Supervised & \underline{0.789} & 0.668 & 0.555 & 0.677 \\
CNN-Aug \cite{wang2020cnn} &  CVPR-20& ResNet50\cite{he2016deep} & Supervised & 0.742 & 0.702 & 0.598 &  0.681\\

FaceX-ray \cite{li2020facexray}  & CVPR-20 & HRNet \cite{li2020facexray} & Supervised & 0.709 & 0.678 & 0.655 & 0.681 \\
FFD \cite{dang2020detection}   & CVPR-20 & Xception\cite{chollet2017xception} & Supervised & 0.784 &0.7435 & 0.605  &  0.711\\
  F$^3$-Net \cite{qian2020thinking}& ECCV-20 & Xception\cite{chollet2017xception} & Supervised & 0.776 & 0.735 &  0.591&  0.700\\
  % SPSL \cite{liu2021spatial}& CVPR-21 & Xception \cite{chollet2017xception} &Supervised  & 0.815 & 0.765 & 0.643 &  0.741\\
  SRM \cite{luo2021generalizing}& CVPR-21 &  Xception \cite{chollet2017xception}& Supervised & \bf 0.792 & \underline{0.755} &  0.601&  0.716\\
  CORE \cite{ni2022core} & CVPR-22 & Xception \cite{chollet2017xception} & Supervised & 0.779 & 0.743 &  0.603&  0.708\\
   RECCE \cite{cao2022end}& CVPR-22 &   Custom CNN & Supervised & 0.767 & 0.731 & 0.609 & 0.702 \\
   UCF \cite{yan2023ucf}& ICCV-23 &  Xception \cite{chollet2017xception} & Supervised & 0.779 & 0.752 & 0.646 &  0.725\\
\hline
\rowcolor{lime} CLIP \cite{ojha2023towards} & CVPR-23& ViT \cite{dosovitskiy2020vit} & Self-Supervised & 0.743 & 0.750 & \underline{0.730} & \underline{0.741} \\

\rowcolor{pink}  Wavelet-CLIP (ours) & - & ViT \cite{dosovitskiy2020vit} & Self-Supervised & 0.756 & \bf 0.759 &\bf 0.732 & \bf 0.749 \\
\hline
\end{tabular}
}
\caption{\textbf{Cross-Data Performance:} The Performance of proposed Wavelet-CLIP with existing state-of-the-art (SOTA) methods using AUC metric ($\uparrow$: more the better). All the supervised models are trained end-to-end on Face Forencics++ \cite{rossler2019faceforensics++} c23 and self-supervised methods are only trained on classification head.  }\label{tab-1}
\end{table*}
\paragraph{Baselines}
This study utilizes the state-of-the-art methods as baselines detailed by Yan \emph{et al. }\cite{yan2024deepfakebench} for evaluating the performance of Wavelet-CLIP. These methods represent the current standard approaches for deepfake detection and are widely recognized in the research community. Broadly, the methods can be categorized into three types: naïve detectors, which employ traditional convolutional neural networks (CNNs) for direct classification; spatial detectors, which explore spatial artifacts or forgery regions in images; and frequency detectors, which analyze manipulation clues in the frequency domain. By using these standard methods—such as Xception \cite{chollet2017xception}, Capsule \cite{nguyen2019capsule}, F3Net \cite{qian2020thinking}—as baseline models, I aim to ensure fair and transparent performance comparisons. Additionally, these baselines serve as a benchmark to assess both cross-domain generalization (using datasets like Celeb-DF v2 and FaceShifter) and unseen deepfake generalization (e.g., testing on large-scale datasets with 50k unseen samples). This approach ensures a robust comparison while highlighting improvements introduced by our Wavelet-CLIP.

Additionally, we reproduce the method Ojha \emph{et al. }\cite{ojha2023towards} which was similar to our approach. Ours and Ojha \emph{et al. }\cite{ojha2023towards} uses a pre-trained self-supervised encoder and do not train or fine-tune it. Their major limitation is that they allow training a Linear Classifier for individual generative  model (DDPM, DDIM and LDM) and the classification head lacks the generalizability. Thus, both these self-supervised encoders are not trained (frozen encoder) and only the classification heads are trained. 

\section{Results}
In this section, we will discuss the performance of our approach with various state-of-the-art approaches.

\paragraph{Cross-Data Performance}
Table \ref{tab-1} clearly demonstrates the superior performance of Wavelet-CLIP compared to existing models. Among the supervised models, SRM \cite{luo2021generalizing} achieves the highest performance on Celeb-DF v1 (0.792) and Celeb-DF v2 (0.755), while other Xception-based methods, such as CORE \cite{ni2022core} and UCF \cite{yan2023ucf}, exhibit competitive but slightly lower results. Notably, methods like MesoNet \cite{afchar2018mesonet} and FFD \cite{dang2020detection} show limited generalization ability, achieving AUCs in the range of 0.609–0.711 on the datasets. Despite being trained end-to-end, these models rely heavily on backbone architectures like Xception \cite{chollet2017xception} and ResNet50 \cite{he2016deep}, which may struggle to generalize under cross-domain settings due to their reliance on supervised learning and dataset-specific artifacts. The self-supervised models CLIP and Wavelet-CLIP demonstrate superior cross-dataset performance, indicating their robustness to unseen data. Unlike the supervised approaches, which require extensive training on specific datasets, self-supervised methods leverage pre-training on large-scale data, enabling better generalization to diverse domains.

%%%%%%%%%%%%%%%%%%%% Table 2 %%%%%%%%%%%%%%%%%%%%%%%%%%%%%%%%%%%%%%%

\begin{table*}[h!]
\centering
\scalebox{1.15}{
\begin{tabular}{ l|cc|cc|cc|cc }
\hline
\rowcolor{lightgray} \multirow{2}{4em}{} & \multicolumn{2}{c|}{\textbf{DDPM} \cite{ho2020denoising}} & \multicolumn{2}{c|}{\textbf{DDIM} \cite{song2020denoising}} & \multicolumn{2}{c|}{\textbf{LDM} \cite{rombach2022high}} & \multicolumn{2}{c}{\textbf{Avg.}} \\ \cline{2-9}
\rowcolor{lightgray}\textbf{Models}&  \textbf{AUC} &  \textbf{EER} & \textbf{AUC} & \textbf{EER} & \textbf{AUC} & \textbf{EER} & \textbf{AUC} & \textbf{EER} \\
\hline
 Xception & 0.712 & 0.353 & 0.729 & 0.331 & 0.658 & 0.309 & 0.699 & 0.331 \\
 CapsuleNet & 0.746 & 0.314 & 0.780 & 0.288 & 0.777 & 0.289 & 0.768 & 0.297 \\
 Core & 0.584 & 0.453 & 0.630 & 0.417 & 0.540 & 0.479 & 0.585 & 0.450 \\
 F$^3$-Net & 0.388 & 0.592 & 0.423 & 0.570 & 0.348 & 0.624 & 0.386 & 0.595 \\
 MesoNet & 0.618 & 0.416 & 0.563 & 0.465 & 0.666 & 0.377 & 0.615 & 0.419 \\
 RECCE & 0.549 & 0.471 & 0.570 & 0.463 & 0.421 & 0.564 & 0.513 & 0.499 \\
 SRM & 0.650 & 0.393 & 0.667 & 0.385 & 0.637 & 0.397 & 0.651 & 0.392 \\
 FFD & 0.697 & 0.359 & 0.703 & 0.354 & 0.539 & 0.466 & 0.646 & 0.393 \\
 MesoInception & 0.664 & 0.372 & 0.709 & 0.339 & 0.684 & 0.353 & 0.686 & 0.355 \\
 SPSL & 0.735 & 0.320 & 0.748 & 0.314 & 0.550 & 0.481 & 0.677 & 0.372 \\
 \hline
 \rowcolor{lime}
 CLIP & 0.781 & 0.292 & 0.879 & 0.203 & 0.876 & 0.210 & 0.845 & 0.235 \\ \hline
 \rowcolor{pink}
Wavelet-CLIP & \bf 0.792 & \bf 0.282 & \bf 0.886 & \bf 0.197 & \bf 0.897 & \bf 0.190 & \bf 0.858 &\bf 0.223 \\
\hline
\end{tabular}
}
\caption{\textbf{Robustness to Unseen Deepfakes:} The Performance of proposed Wavelet-CLIP with existing state-of-the-art (SOTA) methods using AUC ($\uparrow$: more the better) and EER ($\downarrow$: less the better) metrics respectively. All the supervised models are trained end-to-end on Face Forencics++ \cite{rossler2019faceforensics++} c23 and self-supervised methods are only trained on classification head. }\label{tab-2}
\end{table*}

Specifically, for the CDFv1 dataset, traditional CLIP features fall short in capturing detailed representations; a standard ViT-L model \cite{ojha2023towards} achieves an AUC of 0.743, while our model shows a significant improvement with a \(+1.3\%\) increase. In every other scenario, our Wavelet-CLIP model stands out by consistently delivering strong performance. Notably, transformer-based models, including ours and those developed by Ojha \cite{ojha2023towards}, demonstrate effective representation capturing abilities for the FaceShifter (Fsh) dataset \cite{li2020fsh}. Interestingly, the best non-transformer model, Face X-ray \cite{li2020facexray}, achieves an AUC of 0.655, whereas our approach exhibits a significant improvement with a \(+7.7\%\) increase in AUC. The Fsh dataset, known for its sophisticated face manipulation techniques across diverse scenarios, presents a substantial challenge; yet, it appears that pre-trained transformers are particularly adept at discerning the subtle forgeries inherent in such deepfakes.

\paragraph{Robustness to Unseen Deepfakes}
Next, Table \ref{tab-2} assesses the performance of Wavelet-CLIP on face images generated by unseen diffusion-based models. Among the supervised models, CapsuleNet \cite{nguyen2019capsule} and SRM \cite{luo2021generalizing} stand out as strong performers, achieving average AUC scores of 0.768 and 0.651, respectively, with lower EER values compared to other supervised methods like MesoNet \cite{afchar2018mesonet} and FFD \cite{dang2020detection}. However, methods such as Core \cite{ni2022core} and F$^3$-Net \cite{qian2020thinking} exhibit significantly lower performance, with AUC values below 0.6, indicating their poor generalization capability when faced with unseen deepfake types. The CLIP model, which employs a self-supervised learning approach and uses the Vision Transformer (ViT) backbone, significantly outperforms the supervised methods. CLIP achieves an average AUC of 0.845 and a low EER of 0.235, demonstrating its robustness across all three datasets. This highlights the advantage of self-supervised learning in enhancing generalization to unseen data, particularly when compared to traditional supervised methods. Specifically, Wavelet-CLIP achieves the best performance on Celeb-DF v2 (0.759) and FaceShifter (0.732), and competitive results on Celeb-DF v1 (0.756). Compared to CLIP \cite{ojha2023towards}, which also employs a ViT-based backbone, Wavelet-CLIP consistently achieves higher AUC scores by leveraging wavelet-based features to capture both spatial and frequency domain artifacts.  

The aggregated performance of Wavelet-CLIP outperforms standard CLIP \cite{ojha2023towards}, showing an \(+1.3\%\) increase in AUC and a \(-1.2\%\) reduction in EER respectively. This highlights the substantial impact of integrating wavelet transformations within the classification head. Additionally, among the non-transformer based models, Capsule \cite{nguyen2019capsule}—noted for its rotational invariance capabilities—performs best, yet it still falls short of matching Wavelet-CLIP, with performance differences of \(12.5\%\) in AUC and \(10.5\%\) in EER respectively. This further underscores the superior effectiveness of Wavelet-CLIP in handling sophisticated generative challenges. The incorporation of wavelet transforms enhances the model's ability to detect subtle manipulations in images, such as high-frequency discrepancies often introduced by deepfake methods.

% Previously, Ojha \emph{et al. }\cite{ojha2023towards} attempted to use features from ViT L trained in CLIP fashion. Next, utilize a k-NN or train a Linear Classifier (LC)  to classify the generated images sampled from each of the generative model. Their major limitation is that they allow training a Linear Classifier for each generative  model and the classification head lacks the generalizability.

% Thus their major limitation is that they allow training a LC for each generative  model and the classification head lacks the generalizability. The protocol followed by Yan \emph{et al.} \cite{yan2024deepfakebench} could be more appropriate, i.e., all the designed models are trained on on FaceForensics++ dataset \cite{rossler2019faceforensics++} c23. These trained models are tested on the various other datasets for cross-domain evaluation. This evaluation provides good generalization potential to assess model performance. 
% Although, their evaluation does not include testing on novel diffusion models, we study the generalizability  

% But, these models are not tested  for the state-of-the-art diffusion models which have potential to generate photo-relaistic deepfakes. 

%%%%%%%%%%%%%%%%%%%% ICASSP 2025 %%%%%%%%%%%%%%%%%%%%%%%%%%%%%%%

\section{Discussion}
The proposed Wavelet-CLIP framework introduces a novel approach to deepfake detection, combining wavelet-based frequency analysis with features derived from the Vision Transformer (ViT-L/14) pre-trained in a CLIP fashion \cite{radford2021clip}. The extensive experimental evaluations demonstrate that this integration significantly enhances the generalizability and robustness of the detection model, particularly in cross-domain and unseen deepfake scenarios. Unlike traditional supervised methods that rely on dataset-specific artifacts, Wavelet-CLIP leverages self-supervised representations to generalize across diverse generative families, setting a new benchmark for deepfake detection. 

The Wavelet-CLIP is a task-agnostic feature extraction enabled by the CLIP-ViT encoder. Pre-trained on large-scale internet data, the frozen encoder provides strong transferable representations, allowing Wavelet-CLIP to excel across various datasets without the need for task-specific fine-tuning. Additionally, the integration of \textit{wavelet transforms} enables the model to capture fine-grained frequency domain details, complementing spatial features and addressing a key limitation of previous methods like F$^3$-Net \cite{qian2020thinking}. However, the inclusion of wavelet decomposition and reconstruction introduces some computational overhead, which may limit real-time deployment in latency-sensitive applications. Moreover, while CLIP-derived features offer broad generalization, their performance might still depend on the diversity of the pre-training data, potentially limiting their applicability to niche or highly specialized deepfake artifacts. As observed, there is a noticeable improvement in AUC scores, and the model consistently achieves a significant edge in performance for all unseen deepfakes (Refer to Figure \ref{fig-roc-combined}). In Figure \ref{fig-roc-combined}, for certain cases, some models fall below the guessing threshold (AUC = 0.5) on the AUC curve, highlighting their limitations. In contrast, our approach demonstrates a clear advantage, with frozen CLIP-based features already showing a significant edge in detecting unseen facial deepfakes. However, \textit{Wavelet-CLIP} further outperforms the existing state-of-the-art, establishing itself as a standalone solution.

Our current approach is specifically designed for detecting \textbf{facial deepfake images}, and it does not yet address other complex modalities such as audio-based, video-based, or audio-visual deepfakes, which remain significant challenges in the domain. Expanding our method to handle these modalities is crucial, as they often exhibit multi-modal inconsistencies that can be exploited for detection. Additionally, while our approach focuses on fine-tuning the wavelet-based classification head using pre-trained CLIP features, it does not leverage large-scale training on millions of fake images. Such large-scale training could further enhance the model’s ability to capture intricate forged features and improve detection performance. Consequently, our current work is limited in this regard. As part of future research, we identify these limitations as opportunities and stepping stones toward building more comprehensive, multi-modal, and robust deepfake detection systems capable of tackling emerging and diverse manipulation techniques. As a future work, we see that, incorporation of multimodal cues for deepfake detection could leverage models performance \cite{raza2023multimodaltrace, yang2023avoid}. Current detection frameworks rely exclusively on visual features; however, deepfakes often introduce inconsistencies in other modalities, such as audio, speech patterns, or facial expressions in video. For example, combining audio embeddings with visual representations could help detect deepfake videos where audio lip-sync mismatches are present. A multimodal CLIP approach, integrating both visual and auditory signals within a unified feature space, could represent the next step in building robust detection systems.

\section{Conclusions}
Thus, we anticipate a pivotal role for large transformer models, given their proficient ability to discern subtle distinctions by capturing specific nuances from forged features. Overall, Wavelet-CLIP secures state-of-the-art results in cross-data generalization and successfully identifies potential deepfakes originating from diffusion models. As a future direction, we plan to explore the capabilities of large pre-trained transformers on various text guidance-based \cite{rombach2022high}, editing-based\cite{meng2021sdedit}, and translation-based\cite{zhou2024denoising} diffusion models. Such research will establish a foundation for designing detection models capable of thwarting generated deepfakes, even when there is a slight shift in the distribution of the original source.

\section{Reproducibility and Ethics Statement}
To ensure the reproducibility of our results and facilitate further research, we provide complete access to our codebase, pre-trained models, and evaluation protocols. All experiments have been conducted using publicly available datasets, adhering to their respective licensing agreements. The code, along with relevant scripts for data preprocessing and model evaluation, has been made publicly available at \textcolor{red}{\href{https://github.com/lalithbharadwajbaru/wavelet-clip}{link}}. Our work addresses the critical challenge of detecting synthetic images generated by advanced diffusion and autoregressive models, to mitigate the potential misuse of generative AI technologies. While these models have transformative applications in creative domains, they pose risks, including disinformation, privacy violations, and malicious impersonation. Our work is intended solely for research and defensive purposes, such as improving fake image detection systems and enhancing digital media integrity. We acknowledge that detection tools can be exploited to identify weaknesses in generative models for adversarial purposes. 

{\small
\bibliographystyle{ieee_fullname}
\bibliography{egbib}
}

\end{document}